# On Evaluation of Bangla Word Analogies

**Mousumi Akter, Souvika Sarkar, Shubhra Kanti Karmaker ("Santu")**
Big Data Intelligence (BDI) Lab, Auburn University, Alabama, USA
{mza0170, szs0239, sks0086}@auburn.edu

## Abstract

This paper presents a high-quality dataset for evaluating the quality of Bangla word embeddings, which is a fundamental task in the field of Natural Language Processing (NLP). Despite being the 7[th] most spoken language in the world, Bangla is a low-resource language and popular NLP models fail to perform well. Developing a reliable evaluation test set for Bangla word embeddings are crucial for benchmarking and guiding future research. We provide a Mikolov-style word-analogy evaluation set specifically for Bangla, with a sample size of 16678, as well as a translated and curated version of the Mikolov dataset, which contains 10594 samples for cross-lingual research. Our experiments with different state-of-the-art embedding models reveal that Bangla has its own unique characteristics, and current embeddings for Bangla still struggle to achieve high accuracy on both datasets. We suggest that future research should focus on training models with larger datasets and considering the unique morphological characteristics of Bangla. This study represents the first step towards building a reliable NLP system for the Bangla language[1].

## 1 Introduction

The Bangla language, with over 300 million native speakers and ranking as the sixth most spoken language in the world, is often regarded as a language with limited resources (Joshi et al., 2020). Despite the breakthrough in the field of Natural Language Processing (NLP), it has been observed that popular NLP models fail to perform well for low-resource languages like Bangla, despite showing human-like performance in high-resource languages like English.

Due to its significance in NLP tasks, there is a growing need for high-quality Bangla word embedding, and several efforts have been made to create word embedding for Bangla by training with a large corpus (Artetxe and Schwenk, 2019; Bhattacharjee et al., 2022a; Feng et al., 2022). There have been very few efforts to create a high-quality evaluation test set for other NLP tasks like Sentiment Analysis, Machine Translation and Summarization (Hasan et al., 2020, 2021; Akil et al., 2022; Bhattacharjee et al., 2022b). Despite the availability of some word analogy evaluation datasets for other languages (Gurevych, 2005; Hassan and Mihalcea, 2009; Sak et al., 2010; Joubarne and Inkpen, 2011; Panchenko et al., 2016), there is a noteworthy lack of datasets for evaluating the efficacy of word embeddings in low-resource languages, such as Bangla.

We emphasize that the evaluation of Bangla word embeddings is a fundamental task that needs to be addressed. Word embeddings has a direct impact on the performance of different NLP tasks. Therefore, creating a high-quality evaluation test set for Bangla word embeddings will enable researchers to benchmark the performance of existing embedding models and guide further research.

In this paper, we present a Mikolov-style (Mikolov et al., 2013) high-quality word-analogy evaluation set specifically for Bangla, with a sample size of 16678. To the best of our knowledge, we are the first to do so. We have also translated and curated Mikolov's dataset for Bangla, resulting in 10594 samples for cross-lingual research. Our test set includes a variety of tasks that evaluate the quality of Bangla word embeddings, such as word similarity and analogy detection. We also provide an analysis of the performance of several state-of-the-art embedding models: Word2Vec (Mikolov et al., 2013), GloVe (Pennington et al., 2014), fast-Text (Bojanowski et al., 2017), LASER (Artetxe and Schwenk, 2019), LaBSE (Feng et al., 2022), bnBERT (Bhattacharjee et al., 2022a),

---
[1]Dataset can be found here: https://paperswithcode.com/dataset/bangla-word-analogy and we plan to publish the code after acceptance.

bnBART (Wolf et al., 2020) along with our and translated Mikolov dataset.

The lack of available datasets for evaluating the quality of Bangla word embeddings is a significant gap in the NLP research community. Experimental results demonstrate that Bangla has its own unique characteristics, and state-of-the-art techniques show very poor performance in both our and translated Mikolov datasets, highlighting the need for further research in low-resource languages like Bangla.

## 2 Dataset

To evaluate the quality of word vectors, a comprehensive test set was developed that contains three types of semantic questions, and nine types of syntactic questions as shown in Table 1. The semantic questions contain word pairs that are similar in meaning, while the syntactic questions involve relationships between words such as pluralization and verb tense. Overall, the comprehensive test set contains 5036 semantic and 11642 syntactic questions, providing a robust evaluation of the quality of word vectors for the Bangla language. The test set was created by first manually generating lists of similar word pairs, and then forming questions by connecting two pairs. For example, we made a list of 7 divisions and 64 districts they belong to and formed 2160 questions by picking two-word pairs between division-district. Following Mikolov's approach (Mikolov et al., 2013), word analogies were created by using the format $word_B - word_A = word_D - word_C$. The goal was to determine the $word_D$ that is similar to $word_C$ in the same way that $word_A$ is similar to $word_B$.

The creation of the test set also took into account the unique characteristics of the Bangla language. For example, the Bangla language has different forms for numbers, including ordinal forms, female forms, and forms with prefixes and suffixes that change the meaning of the word. Additionally, Bangla has colloquial forms that are present in ancient literature, stories, and novels. A total of 2844-word pairs were formed that reflect these unique characteristics of the Bangla language. Furthermore, 3776-word pairs were included in the test set that is unique from Mikolov's word analogy, such as division-district pairs and number pairs with different forms. These additional word pairs further demonstrate the diverse and complex nature of the Bangla language.

Furthermore, we also translated Mikolov's dataset[2] and manually removed English words that do not have Bangla translations. We also removed word pairs that were duplicated in Bangla, such as present participles and plural verbs. This resulted in a dataset of 10594 samples from the original 19544 samples. It is worth noting that the capital-city and currency relationships present in Mikolov's dataset can also be applied to the Bangla language. In summary, while our dataset focused on the linguistic specifics of the Bangla language, Mikolov's dataset was more focused on common words in both Bangla and English. The translation and curation of Mikolov's dataset provide a useful resource for cross-lingual research and analysis.

## 3 Experimental Setup

**Methods:** We evaluated the quality of word vectors learned in both dataset using both traditional and transformer-based models. We utilized various models, including Word2Vec[3], GloVe[4], and fastText[5], to estimate the quality of word representations. Additionally, we employed transformer-based models, such as LASER[6], LaBSE[7], bnBERT[8], and bnBART[9].

For GloVe, we used the Bengali GloVe model that was trained on Bangla Wikipedia[10] and Bangla news articles. Specifically, we utilized the model trained with 39M tokens. Additionally, for Word2Vec, we used the Bengali Word2Vec model trained with Bengali Wikipedia data (Sarker, 2021). We also used fastText, which provides a model for Bangla word embedding, to further evaluate the quality of word vectors.

For the transformer-based models, we used LaBSE, LASER, bnBERT, and bnBART, which provide sentence embeddings. To obtain the word embedding for a particular word, we passed the word to the model and collected the token embeddings for all the tokens of that word, which were then averaged to obtain the word embedding. We created an exhaustive embedding dictionary for 178152 Bengali words and used the word embeddings from that dictionary to perform the word analogy task. Table 2 shows dimensions of different embedding used. Overall, our study provides a comprehensive evaluation of word vectors learned using various traditional and transformer-based models both for our dataset and translated Mikolov dataset.

---

[2]www.fit.vutbr.cz/ imikolov/rnnlm/word-test.v1.txt
[3]huggingface.co/sagorsarker/bangla_word2vec
[4]huggingface.co/sagorsarker/bangla-glove-vectors
[5]github.com/facebookresearch/fastText/
[6]github.com/facebookresearch/LASER
[7]huggingface.co/sentence-transformers/LaBSE
[8]huggingface.co/csebuetnlp/banglabert
[9]github.com/sagorbrur/bntransformer
[10]bn.wikipedia.org/

| Type of Relationship | Sample # | Examples | | | |
|---|---|---|---|---|---|
| | | Word Pair 1 | | Word Pair 2 | |
| Division-District | 2160 | ঢাকা (Dhaka) | শরীয়তপুর (Shariatpur) | রংপুর (Rangpur) | কুড়িগ্রাম (Kurigram) |
| Gender | 1260 | বাবা (Father) | মা (Mother) | চাচা (Uncle) | চাচী (Aunt) |
| Number-ordinal | 380 | এক (One) | প্রথম (First) | পাঁচ (Five) | পঞ্চম (Fifth) |
| Number-date | 930 | এক (One) | পহেলা (First) | দুই (Two) | দোসরা (Second) |
| Number-female | 306 | এক (One) | প্রথমা (First) | দুই (Two) | দ্বিতীয়া (Second) |
| Antonym-adjective | 4692 | দামী (Expensive) | সস্তা (Cheap) | আসক্ত (Addicted) | নিরাসক্ত (Desperate) |
| Antonym-misc. | 3782 | আগমন (Arrival) | প্রস্থান (Departure) | উজান (Upstream) | ভাটি (Downstream) |
| Tense | 144 | চলছি (Continued) | চলছ (Continuing) | চলব (Will continue) | চলবে (Will Continue) |
| Comparative | 552 | শীতল (Cool) | শীতলতর (Cooler) | দীর্ঘ (Long) | দীর্ঘতর (Longer) |
| Superlative | 600 | দীর্ঘ (Long) | দীর্ঘতম (Longest) | নিম্ন (Low) | সর্বনিম্ন (Lowest) |
| Prefix | 66 | জ্ঞান (Knowledge) | বিজ্ঞান (Science) | শেষ (End) | বিশেষ (Special) |
| Suffix | 94 | ভাব (Attitude) | ভাবখানা (Attitude) | ব্যাপার (Matter) | ব্যাপারখানা (Matter) |
| Affix | 38 | রাত্রি (Night) | রাত্রিতে (Night) | হাতি (Elephant) | হাতিতে (Elephant) |
| Plural-noun | 506 | মাঝি (Sailor) | মাঝিরা (Sailors) | ছাত্র (Student) | ছাত্ররা (Students) |
| Plural-object | 72 | রচনা (Composition) | রচনাবলী (Compositions) | নিয়ম (Rule) | নিয়মাবলী (Rules) |
| Colloquial-Standard-noun | 342 | শুষ্ক (Dry) | শুকনো (Dry) | হস্তী (Elephant) | হাতি (Elephant) |
| Colloquial-Standard-pronoun | 110 | তাহাকে (His) | তাকে (His) | তাহার (His) | তার (His) |
| Colloquial-Standard-verb | 462 | হইলাম (Become) | হলাম (Become) | করিবার (Do) | করার (Do) |
| Colloquial-Standard-conjunction | 182 | পূর্বে (Before) | আগে (Before) | অদ্য (Today) | আজ (Today) |

Table 1: Table shows the statistics of the dataset and examples of the semantic and syntactic relationship set. Rows highlighted in cyan represent unique forms for the Bangla language, while rows highlighted in yellow represent somewhat unique forms, with different syntax compared to English.

**Task Description:** Representing words as vectors to measure similarity has been mainly used for languages with established embeddings, leaving it unexplored for low-resource languages, like Bangla. We explore the use of vector space models for measuring similarity in Bangla, using the approach proposed by Mikolov (Mikolov et al., 2013). Specifically, we focus on more complex similarity tasks, such as identifying words similar to a given number in the context of a date. Specifically, we investigate the question of which Bangla words are similar to দুই (Two) in the same way that এক (One) is similar to প্রথমা (First).

We achieve this by computing a vector X as the difference between the vector representation of প্রথমা (First) and এক (One), added to the vector representation of দুই (Two). We then search for the word closest to X in the vector space, measured by cosine distance, and use it as the answer to the question. We discard the input question words during the search process. When the word vectors are properly trained, this method can correctly identify the most similar word, such as দ্বিতীয়া (Second) in this case.

## 4 Result

We examine the impact of embedding dimension on word analogy accuracy, training data variations, and the accuracy for analogies described in our dataset and the translated Mikolov dataset. Our dataset focuses on unique characteristics of the Bengali language, while the translated Mikolov dataset aids in determining model accuracy for cross-lingual words. For our experiments, we strictly matched our dataset for both Top-1(%) and Top-5(%) accuracy. Achieving maximum accuracy was challenging, but not expected.

**Embedding Dimension Variation:** LaBSE provided a 768-dimensional vector that gave better Top-5% accuracy for semantic relationships around 12%, while bnBART's 1024-dimensional vector gave better syntactic accuracy around 25%, as demonstrated in Figure 1. Top-1% accuracy was generally low in all cases, except for bnBART, which achieved around 21% accuracy.

**Training Data Variation:** Word2Vec, GloVe, fastText, and bnBERT were trained with Bengali Wikipedia and crawled bangla news articles. In contrast, LaBSE and LASER were trained with multilingual data, with LaBSE trained on data from 109 languages and LASER trained on data from 97 languages (Feng et al., 2022). bnBART[11], on the other hand, was trained on data for different inference tasks for the Bengali language, such as QA, NER, MASK generation, Translation, and Text generation. Table 3 demonstrates that training data plays a critical role in the quality of learned word embeddings. Hence, LaBSE and bnBART performed well for different semantic and syntactic relationships and were the overall win-

---
[11]github.com/sagorbrur/bntransformer

| Embedding | Word2Vec | GloVe | fastText | LaBSE | bnBERT | LASER | bnBART |
|---|---|---|---|---|---|---|---|
| Dimension | 100 | 100 | 300 | 768 | 768 | 1024 | 1024 |

Table 2: Dimensions of different embedding used

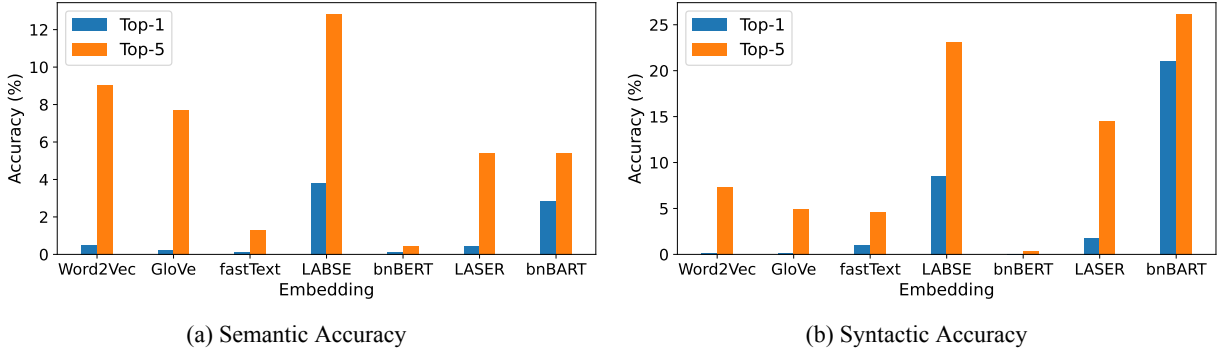

(a) Semantic Accuracy    (b) Syntactic Accuracy

Figure 1: Top-1(%) and Top-5(%) accuracy for different embeddings in our dataset.

| | Embedding | | | | | | |
|---|---|---|---|---|---|---|---|
| **Relationship** | Word2Vec | GloVe | fastText | LaBSE | bnBERT | LASER | bnBART |
| Division-District | **9.8** | 9.5 | 0.0 | 0.1 | 0.6 | 0.0 | 0.0 |
| Gender | 11.2 | 8.5 | 3.3 | **21.9** | 0.1 | 12.7 | 12.6 |
| Number | 5.9 | 5.1 | 0.6 | **16.3** | 0.4 | 3.4 | 3.7 |
| Antonym | 11.0 | 7.6 | **14.5** | 3.5 | 0.7 | 1.0 | 3.3 |
| Tense | 6.3 | 2.8 | 4.2 | 13.9 | 0.0 | **16.7** | 14.6 |
| Comparative | 5.1 | 1.1 | 6.5 | 12.5 | 2.2 | 5.3 | **14.7** |
| Superlative | 3.5 | 1.3 | 3.3 | **10.2** | 0.0 | 6.3 | 5.2 |
| Prefix | 1.5 | 0.0 | 3.0 | 9.1 | 0.0 | 3.0 | **43.9** |
| Suffix | 0.0 | 0.8 | 0.8 | 30.4 | 0.0 | 22.8 | **53.6** |
| Affix | 21.1 | 23.7 | 5.3 | 63.2 | 0.0 | 50.0 | **79.0** |
| Plural | 7.3 | 2.5 | 1.5 | **41.6** | 0.0 | 18.4 | 11.1 |
| Colloquial-Standard | 9.8 | 3.9 | 1.9 | **23.5** | 0.0 | 6.6 | 9.4 |
| **Overall Accuracy** | 7.7 | 5.6 | 3.7 | 20.5 | 0.3 | 12.2 | 20.9 |

Table 3: Top-5 (%) accuracy on three types of semantic and nine types of syntactic relationship set. Highest accuracy for each relationship type of embedding is bold. Results indicate that certain embeddings perform strongly, moderately strong, or weakly depending on the relationship type, with green, blue, and red highlights, respectively.

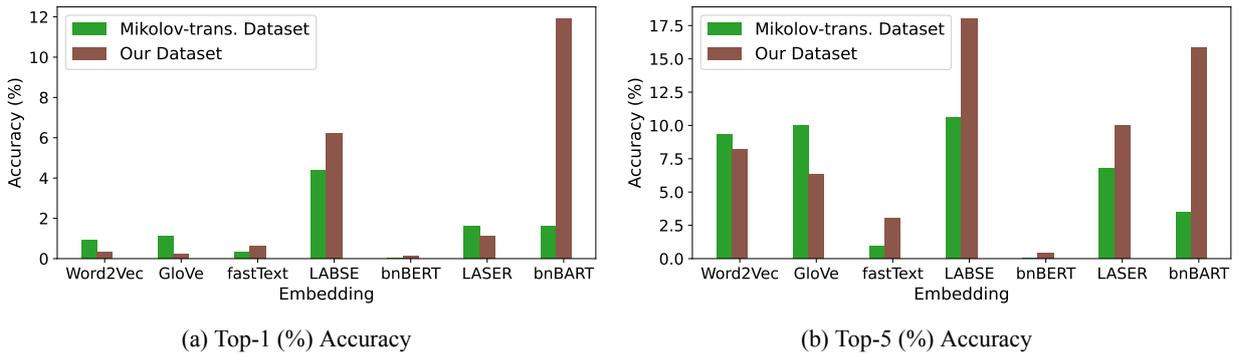

(a) Top-1 (%) Accuracy    (b) Top-5 (%) Accuracy

Figure 2: Comparison of Top-1(%) and Top-5(%) accuracy on translated Mikolov dataset and our dataset.

ners. Task-specific learning appears to aid in the learning of better word embeddings as well.

**Dataset Variation:** We compared our dataset and the translated Mikolov dataset in Figure 2. LaBSE and bnBART were the overall winners again, indicating that more training data including multilingual learning and task-specific learning help learn specific characteristics of the Bangla Language. For more results please look into Appendix 7.

## 5 Conclusion

Despite being the 7th most spoken language[12] in the world, Bangla is still considered a low-resource language in terms of NLP research. Previous studies have attempted to develop NLP tasks for Bangla, but there is a lack of high-quality datasets to evaluate the effectiveness of learned word embeddings. To address this gap, we have presented a high-quality dataset for evaluating Bangla-specific word analogy, similar to the Mikolov dataset. Additionally, we have manually filtered and translated the Mikolov dataset into Bangla, potentially enabling cross-lingual research. Our experiments with different word embeddings suggest that current word embeddings for Bangla still struggle to achieve high accuracy on both datasets. To improve performance, future research should focus on training models with larger datasets and taking into account the unique morphological characteristics of Bangla when developing models for different NLP tasks.

## 6 Limitations

It is important to acknowledge that our study has certain limitations. Firstly, we may have unintentionally dropped some word analogy relations while creating the dataset for Bangla. Therefore, the dataset may not be completely comprehensive, and some relevant relationships might be missing. Additionally, we considered translated dataset from English only, and we did not consider other languages. Therefore, the applicability of our findings to other languages may be limited. Future research can be conducted with the translated dataset from other languages to explore the potential differences in word analogy relations across languages. Furthermore, our study did not explore the impact of other factors such as cultural or contextual differences, which may also influence word analogy relations.

---

[12] en.wikipedia.org/wiki/List_of_languages

# 7 Appendix

|  | Embedding | | | | | | |
| --- | --- | --- | --- | --- | --- | --- | --- |
| **Type of Relationship** | Word2Vec | GloVe | fastText | LaBSE | bnBERT | LASER | bnBART |
| Division-District | **21.5** | 17.5 | 0.0 | 0.1 | 0.7 | 0.0 | 0.0 |
| Gender | 13.3 | 11.6 | 6.4 | **27.5** | 0.1 | 19.4 | 14.3 |
| Number | 9.6 | 9.8 | 1.7 | **21.7** | 1.1 | 5.3 | 5.2 |
| Antonym | 14.6 | 10.5 | **22.6** | 5.3 | 0.8 | 1.6 | 4.7 |
| Tense | 9.7 | 3.5 | 5.6 | 16.0 | 0.0 | **20.1** | 18.8 |
| Comparative | 6.2 | 2.2 | 10.0 | 13.0 | 3.1 | 8.3 | **15.2** |
| Superlative | 5.3 | 1.3 | 5.5 | 12.5 | 0.0 | 8.3 | 5.8 |
| Prefix | 1.5 | 1.5 | 6.1 | 10.6 | 0.0 | 4.6 | **43.9** |
| Suffix | 0.0 | 0.8 | 1.7 | 37.3 | 0.0 | 31.2 | **53.6** |
| Affix | 21.1 | 26.3 | 7.9 | 73.7 | 0.0 | 63.2 | **81.6** |
| Plural | 11.3 | 3.7 | 2.6 | **53.8** | 0.0 | 27.7 | 14.6 |
| Colloquial-Standard | 13.6 | 5.9 | 3.3 | **31.2** | 0.0 | 9.0 | 10.1 |
| **Overall Accuracy** | 10.6 | 7.9 | 6.1 | 25.2 | 0.5 | 16.6 | 22.3 |

Table 4: Top-10 (%) accuracy on three types of semantic and nine types of syntactic relationship set. Highest accuracy for each relationship type of embedding is bold. Results indicate that certain embeddings perform strongly, moderately strong, or weakly depending on the relationship type, with green, blue, and red highlights, respectively.

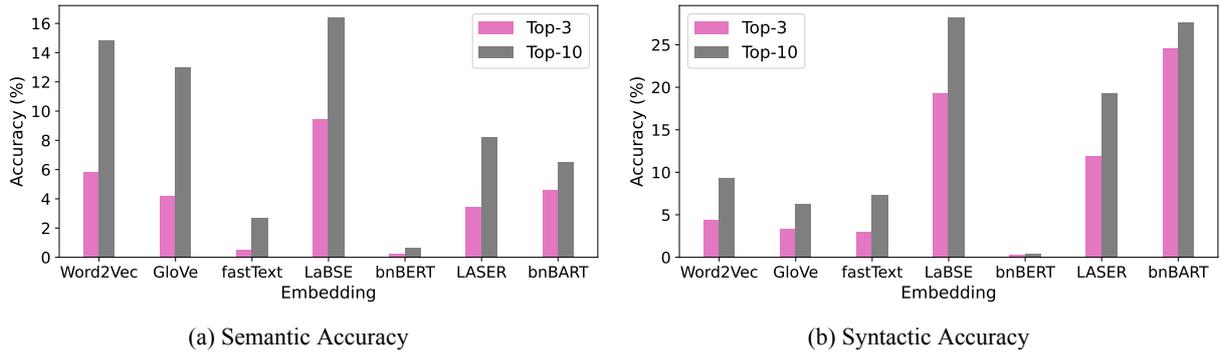

(a) Semantic Accuracy

(b) Syntactic Accuracy

Figure 3: Top-3(%) and Top-10(%) accuracy for different embeddings in our dataset.

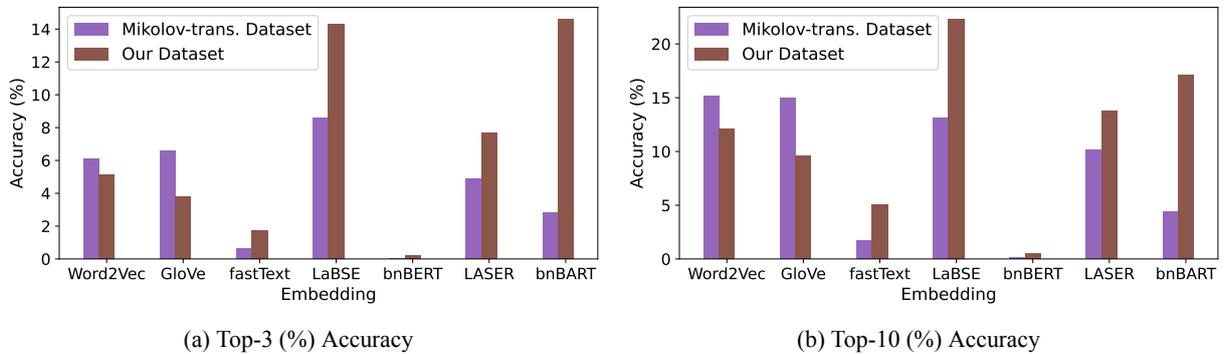

(a) Top-3 (%) Accuracy

(b) Top-10 (%) Accuracy

Figure 4: Comparison of Top-3 (%) and Top-10 (%) accuracy on translated Mikolov dataset and our dataset.